\documentclass{llncs}
\usepackage[numbers]{natbib}

\usepackage{times}
\usepackage{amsmath,amssymb}
\usepackage{pstricks}
\usepackage{graphicx}
\usepackage{xspace}            
\usepackage{graphicx}
\usepackage{multirow}
\usepackage{subfig}
\usepackage{multirow}
\usepackage{array}
\usepackage{url}
\usepackage{pdfpages}
\usepackage{euscript}
\usepackage{mathtools}
\usepackage{enumitem}
\usepackage{epstopdf}
\usepackage{float}
\usepackage[linesnumbered,ruled,vlined]{algorithm2e}
\usepackage{wrapfig}

\newcommand{\vct}[1]{\mathbf #1}

\newcommand{\et}{\textit{et al. }}

\makeatletter
\newif\if@restonecol
\makeatother

\title{Replicator Equation: Applications Revisited}

\author{Tinsae G.Dulecha}
\institute{DAIS, Universit\`a Ca' Foscari di Venezia\\ via Torino 155, 30172 Venezia Mestre, Italy\\
\email{tinsae.dulecha@unive.it}
}

\begin{document}

\maketitle

The replicator equation is a simple model of evolution that leads to stable form of Nash Equilibrium, Evolutionary Stable Strategy (ESS). No individuals get an incentive unilaterally deviating from the equilibrium. It has been studied in connection with Evolutionary Game Theory, a theory John Maynard Smith and  George R. Price developed to predict biological reproductive success of populations. Replicator equation was originally developed for symmetric games, games whose payoff matrix is skew-symmetric and in general tells us, in a game, how individuals or populations change their strategy over time based on the payoff matrix of the game. Consider a large population of players where each player is assigned a strategy and players cannot choose their strategies which means rationality and consciousness don't enter the picture (players play based on a pre-assigned set of strategies).  Evolutionary game theory assumes a scenario where a non-rational pairs of players, which play based on a pre-assigned set of strategies, repeatedly drawn from a large population  plays a symmetric two-player game which drives the strategies with lower payoff to extinction .  Since players interacts with another randomly chosen player in the population, a player’s expected payoff is determined by the payoff matrix and the proportion of each strategy in the population. The limiting behavior of the replicator dynamics (i.e., the evolutionary outcome) are the Evolutionary Stable Strategies, a NE with additional stability properties.

Let $x_i(t)$ is the proportion of the population which plays strategy $i$ $\in N$(set of strategies) at time $t$. The state of the population at any given instant is then given by $\vct{x}_i(t)$ = [$x_1,x_2, ..., x_n$]$'$ where $'$ denotes transposition and $n$ refers the size of available pure strategies, $|N|$. With $A$ be the $n \times n$ payoff matrix ( biologically measured as Darwinian fitness i.e reproductive success ), the payoff for the $i^{th}$-strategist, assuming the opponent is playing the $j^{th}$ strategy, is $(a_{ij})$, the corresponding $i^{th}$ row and the $j^{th}$ column of $A$. If the population is in state x, the expected payoff earned by an the $i^{th}$-strategist is $(A\vct{x})_i$ while the mean payoff over the whole population is $\vct{x}'A\vct{x}$.
The growth rate of the population is then computed as the product of the current frequency
of strategy with the own fitness relative to the average (the difference in the expected and the average payoff gives us the per capita rate). The game, which is assumed to be played over and over, generation after generation, changes the state of the population over time until equilibrium is reached, a point $\vct{x}$ is said to be a stationary (or equilibrium) point of the dynamical system if $\dot{x}$ = 0 where the dot implies derivative. Different formalization of this selection process have been proposed in evolutionary game theory where replicator dynamics is one of the well knowns. 

Beyond its first emphasis in biological use, evolutionary game theory has been expanded well beyond in social studies for behavioral analysis, in machine learning, computer vision and others. Its application in the fields of computer science has drawn my attention which is the reason to write this extended abstract.

Torsello \et has shown that dominant sets, a well known clustering notion which generalizes the maximal clique problem to edge weighted graphs, can be bijectively related to Evolutionary Stable Strategies (ESS) \cite{TorPellRot06}. Since the formalization of clustering problem as a game, replicator dynamics has been used to address different problems \cite{TorPellRot06}. Players simultaneously select an object to cluster and receive a payoff proportional to the similarity of the selected objects. Since clusters are sets of objects that are highly similar, the competition induced by the game forces the players to coordinate by selecting objects from the same cluster. Indeed, by doing so, both players are able to maximize their expected payoff. The clustering game is a non-cooperative game between two-players, where the objects to cluster form the set of strategies, while the affinity matrix provides the players’ payoffs. The dynamics is also able to solve multiple affinities by extending the clustering game, which has a single payoff, to a multi-payoff game \cite{MeqBulPelSIMBAD2015}. Since the introduction of the connection, replicator equation helps dominant set formulation be a robust clustering approach with many interesting and powerful properties such as: it makes no assumption on the underlying data representation, it makes no assumption on the structure of the affinity matrix, being it able to work with asymmetric and even negative similarity functions alike; it does not require a priori knowledge on the number of clusters; able to solve one-class clustering problems (figure ground separation); it allows extracting overlapping clusters \cite{TorBulPelICPR2008}; it generalizes naturally to hypergraph clustering problems, in which the clustering game is played by more than two players \cite{RotPel13}. The dominant sets formulation, using replicator equation, have proven to be an effective tool for graph-based clustering and have found applications in a variety of different domains, such as bioinformatics \cite{bioinformatics3}, computer vision \cite{computervision6}, image processing \cite{ItSecurity},\cite{matching}, group detection \cite{fFormation}\cite{groupDetection1}\cite{SebEyaCVIU2016}, security and video surveillance  \cite{securityAndVideo}, \cite{securityAndVideo1}, \cite{YonEyaPelPraIET2016}, \cite{HamMelAndPel} etc. Very recently, Eyasu $\et$, in their two consecutive works \cite{MeqBulPelSIMBAD2015,ZemPelECCV16}, generalized the dominant set formulation. The generalized version, using replicator dynamics as a solver, can be applied in different fields of studies such as in computer vision, biomedical analysis, human behavior and social network analysis and others.

The notion of clustering using replicator equation has also be used in \cite{DonBisCVPR2013} where the core idea is to combine an effective diffusion
process, based on iteratively approaching evolutionary stable strategies. Similarly, Xingwei \et used it for dense neighbor selection for affinity learning with diffusion on tensor product graph. Few of the most efficient and most recent computer vision applications that uses replicator equations include: Retrieval\cite{EyaLeuPeliICPR2016} where replicator is used to different size dense neighbors which improves the quality of affinity propagation; tracking \cite{YonEyaPelPraIET2016} which uses replicator as a solver; interactive image segmentation where the segmentation is guided by the user provided information \cite{ZemPelECCV16}; large scale image geo-localization \cite{EyaYonHarAndMarMubPAMI} 

Morteza showed, in his adaptive trajectory analysis of replicator dynamics, how much effective replicator dynamics is for data clustering and structure identification \cite{MorML2016}. Extraction of dense subgraphs using game dynamics is done in 
\cite{HaiLonShuPAMI2013}, and in \cite{MeqPelICIAP2015} it has been shown that replicator have no difficulty in extracting a few “natural” clusters from heavy background
noise. Moreover, Eyasu \et proposed effective replicator based algorithm for simultaneous clustering and outlier detections \cite{EyaYonAndPeliICPR2016}. 

\bibliographystyle{splncsnat}
\bibliography{MyBibFile}

\begin{thebibliography}{23}
\providecommand{\natexlab}[1]{#1}
\providecommand{\url}[1]{\texttt{#1}}
\providecommand{\urlprefix}{}

\bibitem[{Alata et~al.(2006)Alata, Dacier, Deswarte, Kaa{\^a}niche,
  Kortchinsky, Nicomette, Pham, and Pouget}]{securityAndVideo1}
Alata, E., Dacier, M., Deswarte, Y., Kaa{\^a}niche, M., Kortchinsky, K.,
  Nicomette, V., Pham, V.H., Pouget, F.: Collection and analysis of attack data
  based on honeypots deployed on the internet.
\newblock In: Quality of Protection, pp. 79--91 (2006)

\bibitem[{Albarelli et~al.(2009)Albarelli, Rota~Bul\`o, Torsello, and
  Pelillo}]{matching}
Albarelli, A., Rota~Bul\`o, S., Torsello, A., Pelillo, M.: Matching as a
  non-cooperative game.
\newblock In: ICCV'09. pp. 1319--1326 (2009)

\bibitem[{Chehreghani(2016)}]{MorML2016}
Chehreghani, M.H.: Adaptive trajectory analysis of replicator dynamics for data
  clustering.
\newblock Machine Learning 104(2-3), 271--289 (2016)

\bibitem[{Donoser and Bischof(2013)}]{DonBisCVPR2013}
Donoser, M., Bischof, H.: Diffusion processes for retrieval revisited.
\newblock In: {IEEE}, {CVPR}. pp. 1320--1327 (2013)

\bibitem[{Frommlet(2010)}]{bioinformatics3}
Frommlet, F.: Tag snp selection based on clustering according to dominant sets
  found using replicator dynamics.
\newblock Adv. Data Analysis and Classification 4(1), 65--83 (2010)

\bibitem[{Hamid et~al.(2015)Hamid, Melaku, Pelillo, and Prati}]{HamMelAndPel}
Hamid, A.K., Melaku, L.S., Pelillo, M., Prati, A.: Using dominant sets for data
  association in multi-camera tracking.
\newblock In: Proceedings of the 9th International Conference on Distributed
  Smart Camera, Seville, Spain, September 8-11, 2015. pp. 38--43 (2015)

\bibitem[{Hamid et~al.(2005)Hamid, Johnson, Batta, Bobick, Isbell, and
  Coleman}]{securityAndVideo}
Hamid, R., Johnson, A.Y., Batta, S., Bobick, A.F., Isbell, C.L., Coleman, G.:
  Detection and explanation of anomalous activities: Representing activities as
  bags of event n-grams.
\newblock In: CVPR (1). pp. 1031--1038 (2005)

\bibitem[{Hung and Kr{\"{o}}se(2011)}]{fFormation}
Hung, H., Kr{\"{o}}se, B.J.A.: Detecting f-formations as dominant sets.
\newblock In: Proceedings of the 13th International Conference on Multimodal
  Interfaces, {ICMI} 2011, Alicante, Spain, November 14-18, 2011. pp. 231--238
  (2011)

\bibitem[{Liu et~al.(2013)Liu, Latecki, and Yan}]{HaiLonShuPAMI2013}
Liu, H., Latecki, L.J., Yan, S.: Fast detection of dense subgraphs with
  iterative shrinking and expansion.
\newblock {IEEE} Trans. Pattern Anal. Mach. Intell. 35(9), 2131--2142 (2013)

\bibitem[{Mequanint et~al.(2015)Mequanint, {Rota Bul{\`{o}}}, and
  Pelillo}]{MeqBulPelSIMBAD2015}
Mequanint, E.Z., {Rota Bul{\`{o}}}, S., Pelillo, M.: Dominant-set clustering
  using multiple affinity matrices.
\newblock In: {SIMBAD}. pp. 186--198 (2015)

\bibitem[{{Rota Bul\`o} and Pelillo(2013)}]{RotPel13}
{Rota Bul\`o}, S., Pelillo, M.: A game-theoretic approach to hypergraph
  clustering.
\newblock IEEE Trans. Pattern Anal. Machine Intell. 35(6), 1312--1327 (2013)

\bibitem[{Tesfaye et~al.(2016)Tesfaye, Zemene, Pelillo, and
  Prati}]{YonEyaPelPraIET2016}
Tesfaye, Y.T., Zemene, E., Pelillo, M., Prati, A.: Multi-object tracking using
  dominant sets.
\newblock IET computer vision 10, 289–298 (2016)

\bibitem[{Torsello et~al.(2006)Torsello, {Rota Bul\`o}, and
  Pelillo}]{TorPellRot06}
Torsello, A., {Rota Bul\`o}, S., Pelillo, M.: Grouping with asymmetric
  affinities: A game-theoretic perspective.
\newblock In: Proc. {IEEE} Conf. Computer Vision and Pattern Recognition
  (CVPR). pp. 292--299 (2006)

\bibitem[{Torsello et~al.(2008)Torsello, Bul{\`{o}}, and
  Pelillo}]{TorBulPelICPR2008}
Torsello, A., Bul{\`{o}}, S.R., Pelillo, M.: Beyond partitions: Allowing
  overlapping groups in pairwise clustering.
\newblock In: ICPR. pp. 1--4 (2008)

\bibitem[{Torsello and Pelillo(2009)}]{computervision6}
Torsello, A., Pelillo, M.: Hierarchical pairwise segmentation using dominant
  sets and anisotropic diffusion kernels.
\newblock In: EMMCVPR. pp. 182--192 (2009)

\bibitem[{Vascon et~al.(2016)Vascon, Mequanint, Cristani, Hung, Pelillo, and
  Murino}]{SebEyaCVIU2016}
Vascon, S., Mequanint, E.Z., Cristani, M., Hung, H., Pelillo, M., Murino, V.:
  Detecting conversational groups in images and sequences: {A} robust
  game-theoretic approach.
\newblock Computer Vision and Image Understanding 143, 11--24 (2016)

\bibitem[{Vascon et~al.(2014)Vascon, Zemene, Cristani, Hung, Pelillo, and
  Murino}]{groupDetection1}
Vascon, S., Zemene, E., Cristani, M., Hung, H., Pelillo, M., Murino, V.: A
  game-theoretic probabilistic approach for detecting conversational groups.
\newblock In: ACCV (2014)

\bibitem[{Wang et~al.(2008)Wang, Ye, Wang, and Wang}]{ItSecurity}
Wang, M., Ye, Z.L., Wang, Y., Wang, S.X.: Dominant sets clustering for image
  retrieval.
\newblock Signal Processing pp. 2843--2849 (2008)

\bibitem[{Zemene et~al.(2016{\natexlab{a}})Zemene, Alemu, and
  Pelillo}]{EyaLeuPeliICPR2016}
Zemene, E., Alemu, L.T., Pelillo, M.: Constrained dominant sets for retrieval.
\newblock In: ICPR (2016{\natexlab{a}})

\bibitem[{Zemene and Pelillo(2015)}]{MeqPelICIAP2015}
Zemene, E., Pelillo, M.: Path-based dominant-set clustering.
\newblock In: {ICIAP}. pp. 150--160 (2015)

\bibitem[{Zemene and Pelillo(2016)}]{ZemPelECCV16}
Zemene, E., Pelillo, M.: Interactive image segmentation using constrained
  dominant sets.
\newblock In: {ECCV} 2016. pp. 278--294 (2016)

\bibitem[{Zemene et~al.(2017)Zemene, Tariku, Idrees, Prati, Pelillo, and
  Shah}]{EyaYonHarAndMarMubPAMI}
Zemene, E., Tariku, Y., Idrees, H., Prati, A., Pelillo, M., Shah, M.:
  Large-scale image geo-localization using dominant sets.
\newblock CoRR abs/1702.01238 (2017)

\bibitem[{Zemene et~al.(2016{\natexlab{b}})Zemene, Tariku, Prati, and
  Pelillo}]{EyaYonAndPeliICPR2016}
Zemene, E., Tariku, Y., Prati, A., Pelillo, M.: Simultaneous clustering and
  outlier detection using dominant sets.
\newblock In: ICPR (2016{\natexlab{b}})

\end{thebibliography}
\end{document}